# Optimal Policy Sparsification and Low Rank Decomposition for Deep Reinforcement Learning

Vikram Goddla, Detroit Country Day School, Beverly Hills, MI.

February 2024

## Abstract

Deep reinforcement learning (DRL) has shown significant promise in a wide range of applications including computer games and robotics. Yet, training DRL policies consume extraordinary computing resources resulting in dense policies which are prone to overfitting. Moreover, inference with dense DRL policies limit their practical applications, especially in edge computing. Techniques such as pruning and singular value decomposition have been used with deep learning models to achieve sparsification and model compression to limit overfitting and reduce memory consumption. However, these techniques resulted in sub-optimal performance with notable decay in rewards. $L_1$ and $L_2$ regularization techniques have been proposed for neural network sparsification and sparse auto-encoder development, but their implementation in DRL environments has not been apparent. We propose a novel $L_0$-norm-regularization technique using an optimal sparsity map to sparsify DRL policies and promote their decomposition to a lower rank without decay in rewards. We evaluated our $L_0$-norm-regularization technique across five different environments (Cartpole-v1, Acrobat-v1, LunarLander-v2, SuperMarioBros-7.1.v0 and SurROL-Surgical Robot Learning) using several on-policy and off-policy DRL algorithms. We demonstrated that the $L_0$-norm-regularized DRL policy in the SuperMarioBros environment achieved 93% sparsity and gained 70% compression when subjected to low-rank decomposition, while significantly outperforming the dense policy. Additionally, the $L_0$-norm-regularized DRL policy in the SurROL-Surgical Robot Learning environment achieved a 36% sparsification and gained 46% compression when decomposed to a lower rank, while being performant. The results suggest that our custom $L_0$-norm-regularization technique for sparsification of DRL policies is a promising avenue to reduce computational resources and limit overfitting.

# 1  Introduction

Deep reinforcement learning policies can become excessively large with dense weights at every layer especially when applied to complex tasks. Such large models, with dense weights at every layer are prone to overfitting and tend to yield sub-optimal performance[1]. Parameter sparsification using Pruning techniques[2] have been proposed to reduce the size of deep neural-network-based models for resource efficient inference. Pruning uses a simple procedure to slowly transform a dense network to a sparse network using weight magnitudes[3]. DeltaNetwork algorithms have also been proposed in combination with pruning to achieve temporal sparsity during inference[4]. However, Pruning techniques need a fully trained DRL policy and consequently requiring added effort - first train the DRL policy and then prune the DRL policy. Moreover the success of pruning algorithms vary significantly and often leading to under performance[5][6]. Dynamic Sparse Training methods such as Sparse Evolutionary Training algorithm[7],[8] were introduced to leverage evolutionary techniques to induce sparsity in DRL policies during training, with a random sparse network as a starting point. There have been some findings that reliance on random starting point may lead to local minima and low accuracy[9]. Regularization techniques such as $L_1$ and $L_2$ norm regularization[10] [11] have been proposed to induce sparsity in deep neural-network-based models including sparse auto-encoders. Although such applications in the Deep Reinforcement Learning space have not been apparent. $L_1$ and $L_2$ norm regularization induce sparsity in the deep neural-network-based-models via parameter shrinkage [10]. $L_1$ norm regularization induces sparsity by shrinking some parameters to zero and hence producing relatively sparse models but they can be overly biased leading to poorly fitted models[12]. On the other hand $L_2$ norm regularization tends parameters towards zero (0) and hence requiring thresholds to induce sparsity.

To overcome such challenges, we propose a custom implementation of the $L_0$ norm regu-larization for deep reinforcement learning via a learned sparsification map governed by a sparsification penalty coefficient. Instead of applying parameter shrinkage to an existing dense network, the proposed method aims to achieve sparsity while training the deep reinforcement policy. We demonstrate via various bench-marking studies that our custom implementation of $L_0$ norm regularization results in an optimal sparse policy which yields equivalent (or better) performance without decay in rewards when compared to that of a corresponding dense policy. In addition to being prone to overfitting, deep-neural-network-based policies tend to have large memory consumption on storage devices. Low rank decomposition techniques, such as Singular Value Decomposition ($SVD$), have commonly been used to reduce the amount of memory consumed by such policies. $SVD$ attempts to factorize a given high rank matrix as the product of three low rank matrices which typically consume less memory compared to the original matrix. The dot product of the low rank matrix factors gives an approximate version of the supposedly high rank matrix. Consequently, the low rank approximated policy does tend to be a sub-optimal version of the original matrix. We demonstrate that a sparse policy generated by our custom implementation of $L_0$ norm regularization promotes decomposition to a lower rank, when compared to a dense policy, while achieving equivalent or better performance with higher rewards than those of the dense policy. Previous studies have generally reported reductions in parameter count and floating-point operations (FLOPs)[7],[8]. However, to our knowledge, explicitly demonstrated evidence of sparsification of deep reinforcement learning policies with low rank decomposition using $L_0$ norm regularization, without decay in rewards, has yet to be reported.

## 2   Related Works

Over the years, there have been various studies related to neural network based model/policy sparsification. One, that is similar to our approach, in the context of reinforcement learning, is provided by Li et al. in their study related to the regularized Markov Decision Process (*MDP*) [13]. This study proposes a framework that utilizes entropy regularization to find the optimal policy that maximises the reward function and a regularization term. It also suggests that the proposed framework can bring about multi-modality and policy sparsity via the use of many regularization terms. The major difference between the norm regularization approach we used and the regularized *MDP* approach is the fact that in our approach, a sparsification map is computed via learned "location" parameter in log space $log(\alpha)$, alongside a temperature value $\beta$ that contributes to the sparsification penalty, whereas in the regularised *MDP* technique an entropy-like regularization function is used to induce regularization of policy to minimise the bellman residue for a given Q-value function, which consequently induces sparsification of policy. There is also a more recent literature that proposed a sparsity induced system representation for policy decomposition[14]. This technique depends on heuristic search, to find the best representation of policy, best amenable to decomposition with lower but reasonable suboptimality.

There are other proposed techniques related to the sparsification of neural network based models that are not necessarily directed towards the field of reinforcement learning but can still be adapted to it. Techniques like the lottery ticket hypothesis[15], where a neural network model is hypothesised to contain sub networks with fewer learning parameters that are just as capable of learning the data distribution as the full network. The technique is based on "weight pruning" and works well enough on small/shallow networks, however they do tend to not be as optimal with large/deep neural networks due to various factors, including noise induced by network depth. There is also the widely known and more common technique of parameter neuron pruning[16], that aims to improve generalizability of a neural network based model by removing "unimportant" neurons from the learned parameters, hence reducing model complexity and improving its interpretability, reducing computational time and cost, while still maintaining reasonable accuracy and error. There is also literature that proposes the technique of sparsification of probabilistic models[17] that maintain the ego-betweenness of neurons present in the sparsified models.

Additionally, several studies including Swaminathan et al[18], Hoefler et al.[19] and Elsen et al.[20] have suggested that sparse matrices in network architecture facilitate reduced memory usage and enhanced computational efficiency during inference. Moreover, several studies have suggested that sparse matrices may promote decompostion to a lower rank when compared to dense matrices. For example, the singular value decomposition ($SVD$) of a matrix $A$ can be written as $A = USV^\top$, where $U$ and $V$ are orthogonal matrices, and $S$ is a diagonal matrix containing the singular values sorted in descending order. A low-rank approximation of $A$ can be achieved by retaining only the first $k$ singular values (and corresponding vectors in $U$ and $V$), especially when these values significantly dwarf the subsequent ones. This is an optimal approximation in the Frobenius norm sense [21, 22]. The choice of $k$ is crucial; a suitable choice captures the most significant components of the original matrix while significantly reducing its rank, which is especially advantageous for sparse matrices. Notably, the truncated $SVD$ offers lossy compression, reducing storage space requirements from $O(mn)$ to $O(k(m+n))$. Such advantages have been noted in various studies, including but not limited to the work of Golub and Van Loan in Ref. [21, 22]. on matrix computations and Halko et al.) on finding structure with randomness[21, 23].

## 3 Preliminaries

### 3.1 *DQN* Technique

*DQN* [24] is a variant of the traditional Q-Learning [25] that utilizes neural networks to approximate the optimal policy rather than storing said optimal policy in a table. Due to the emergence of neural networks and the limitations of the traditional Q-Learning with more complex observation and action spaces, *DQN* was formulated. *DQN* is an off-policy reinforcement learning approach that utilizes a neural network agent and target policy to maximize the total cumulative reward from the environment by taking optimal actions. This approach, just like the traditional Q-Learning, utilizes explorative and exploitative techniques in order to learn the best optimal policy for the given environment. The explorative and exploitative nature of the policy during training is governed by the epsilon greedy strategy. This strategy utilises an epsilon value $\epsilon$ that starts out as 1, making the agent more explorative than exploitative and gradually decays to 0 via cosine annealing, making the agent more exploitative than explorative. As the agent explores the environment, data pertaining to the state $s_t$, the action taken at that state $a_t$, the reward of the action $r_t$, the next state $s_{t+1}$, and the terminal state $d$ are collected and stored in "Replay Memory." Over time batches of data are sampled from the replay memory to update the neural network via a stochastic gradient descent technique. Prior to training, the agent and the target networks are initialized. The target network is a carbon copy of the agent network and is initialized with the same network parameters as the agent. The agent network is made to take in various states (observations) and output Q values pertaining to possible actions at those given states. During training, the agent takes actions based on the value of $\epsilon$. As $\epsilon \to 1$, the agent takes a more explorative action $a_t \sim U\{a_1, a_2, \cdots, a_n\}$, where $U$ is a uniform distribution with all possible actions in its space. As $\epsilon \to 0$, the agent takes a more exploitative action $a_t = argmax(\pi_\theta(s_t))$, where $\pi_\theta$ is the agent policy. For each action taken, a sample record of the experience $e_t = s_t, a_t, r_t, s_{t+1}, d$ is stored in the replay buffer. When the size of the replay buffer has reached a certain amount, usually equal to or more than the batch size, the parameters of the agent policy are updated via stochastic gradient descent.

In the agent policy parameter update, a batch of $s_t$ is fed to the agent policy $\pi_\theta$ to output $Q$ values $\pi_\theta(s_t)$ and a batch of $s_{t+1}$ is fed to the target policy $\bar{\pi}_\theta$ to output future $Q$ values $\bar{\pi}_\theta(s_{t+1})$. Next, the average square difference between the batch of agent $Q$ values corresponding to the actions $a_t$ and the batch of true optimal $Q$ values is computed with the following expression:

$$L_\theta = \left\| Q(s_t)_{[a_t]} - (r_t + (\gamma \cdot max(\hat{Q}(s_{t+1})_{[a_{t+1}]}) \cdot (1-d))) \right\|^2 \quad (2.1)$$

where: $\gamma$ is the discount factor for future rewards.

We can observe the Bellman equation [26] in 2.1, used to compute the batch of true optimal Q values. The agent policy parameters are updated based on the gradient of the loss term $L_\theta$. The target policy network is not updated in the same manner as the agent policy, but rather it is updated at certain intervals. The update frequency is typically defined by a hyper parameter $C$. Periodically, the target policy network is simply updated by replacing its weights with that of the agent policy network. By doing so we can avoid the target Q values being highly correlated with the agent Q values, which can lead to instability in training.

## 3.2 *DDQN* Technique

*DDQN* is a variant of *DQN* that aims to minimize overestimation bias[24]. Overestimation bias is a phenomenon where the policy estimates a *Q-value* greater than the target *Q-value*, thus leading to a negative temporal difference value. Overestimation bias can lead to bias sampling of experiences from the replay memory via their corresponding priority experiences. With the classic *DQN*, for a given experience $e_t = s_t, a_t, r_t, s_{t+1}, d$, the temporal difference $TD_t$ is given as:

$$TD_t = (r_t + (\gamma \cdot \max(\bar{\pi}\theta(s_{t+1}, a_t)) \cdot d)) - \pi_\theta(s_t, a_t) \quad (2.2)$$

The priority of this experience an be computed using the following:

$$P_{exp} = TD_t^{\alpha} \quad where \;\; 0 \leq \alpha \leq \infty \qquad (2.3)$$

With the *DDQN* approach the temporal difference $TD_t$ for experience $e_t$ is given as:

$$TD_t = (r_t + (\gamma \cdot \max(\bar{\pi}\theta(s_{t+1}, argmax(\pi_\theta(s_t, a_t)))) \cdot d)) - \pi_\theta(s_t, a_t) \qquad (2.4)$$

Once the experience priority is computed for all experience records in the buffer memory, one can compute the mutually exclusive probabilities for each experience via the softmax operation on all experiences available in the buffer. By doing so, one can generate the probability values corresponding to the chances of experiences being sampled from the buffer memory and use them to train the policy.

It is quite intuitive to understand why computing experience priorities can be helpful. The temporal difference is an indication of how good an estimated *Q-value* is in comparison to its corresponding target *Q-value*, and the priority of experiences from the equations 2.2 and 2.4 above are also directly proportional to their corresponding temporal differences. In other words the more the temporal difference, the more its priority thus the reason it will be sampled more times than ones with lesser priority. Another interpretation to this is that experiences with higher priorities happen to be more difficult to model as they contain more information than those with lower priorities, and thus are sampled more.

## 3.3 *PPO* Technique

Unlike the *DQN* approach, *PPO* [27] is an on-policy technique, as it collects minibatches of experience while interacting with the environment and uses them to update its policy, whereas off-policy models use different functions (policies) to collect experiences than the one that is being updated for learning [28]. *PPO* aims to maximize the objective:

$$L_\theta = \hat{E}_t \left[ \frac{\pi_\theta(s_t \mid a_t)}{\pi_{\theta_{old}}(a_t \mid s_t)} \cdot \hat{A}_t \right] \quad (3.1)$$

Subject to: $\quad \hat{E}_t \left[ KLD(\pi_{\theta_{old}}(a_t \mid s_t), \pi_\theta(s_t \mid a)) \right] \leq \delta \quad (3.2)$ where: $\hat{E}_t$ corresponds to the expected value, $\pi_{\theta_{old}}(a_t \mid s_t)$ and $\pi_{\theta_{new}}(a_t \mid s_t)$ correspond to the old and new policy estimates and $\hat{A}_t$ corresponds to the Generalized Advantage Estimation($GAE$), expressed as:

$$\hat{A}_t = \sum_{i=t}^{T-1} (\gamma \cdot \lambda)^{i-t} \cdot \hat{\delta}_i \quad (3.3)$$

$$\hat{\delta}_t = r_t + \gamma \cdot \pi_{\theta V}(s_{t+1}) - \pi_{\theta V}(s_t) \quad (3.4)$$

Where: $r_t$ is the reward at time $t$, $\gamma$ is the discount factor, $\hat{\delta}_t$ is the discounted return at time $t$, $\lambda$ is the decay factor for prior returns to the return at time $t$, and $\pi_{\theta V}$ is the value model responsible for estimating the value return of a state at any given state in time.

Let:

$$r_t(\theta) = \frac{\pi_\theta(a_t \mid s_t)}{\pi_{\theta_{old}}(a_t \mid s_t)} \quad (3.5)$$

Then:

$$L_\theta = \hat{E}_t \left[ r_t(\theta) \cdot \hat{A}_t \right] \quad (3.6)$$

It can be observed that (3.1) is a constrained optimization problem as it is subject to a constraint which is (3.2). Constrained Optimization problems are not particularly as easy to solve as unconstrained ones, because they require a conjugate gradient which at times can be more expensive to compute and propagate. In *PPO*, we aim to maximize the objective:

$$L_\theta = \hat{E}_t \left[ min(r_t(\theta) \cdot \hat{A}_t, clip(r_t(\theta), 1 - \epsilon, 1 + \epsilon) \cdot \hat{A}_t) \right] \quad (3.7)$$

*Note: Maximizing the objective $L_\theta$ is equivalent to minimizing $-L_\theta$.*

The first term inside the *min* is the *CPI* (Conservative Policy Iteration) objective. Using this objective alone, the policy model may not converge as its current policy would tend to deviate largely from the old policy. To ensure convergence, the *PPO* utilizes a clipped surrogate function to the CPI function, governed by an $\epsilon$ hyperparameter. By doing so, the model is certain to converge, while also simultaneously ensuring that the new policy is in close proximity to the old policy, thus the name "Proximal Policy Optimization". In our implementation, we used an $\epsilon$ value of 0.2 and also added an "entropy bonus" as suggested by Schulman et al.[27], to maximize entropy and increase exploration.

## 3.4 DDPG with HER and DEX

### 3.4.1 DDPG Technique

The DDPG (Deep Deterministic Policy Gradient) technique is an off-policy technique which is comprised of an actor and a critic network which is effective in continuous control environments[29]. The actor network is responsible for taking action(s) for a given state. The critic network is responsible for estimating the Q values of the state and action pairs of the actor. In this algorithm, the aim of the actor is to maximise the reward which effectively maximises the estimated Q value of the critic and the aim of the critic is to minimise the discrepancy between the estimated value and the target Q value, as outlined below:

$$\max(\mathbb{E}[V_\theta(S_t, a_t)]) \ (3.8); \quad \min([V_\theta(s_t, a_t)\text{-}(r + \gamma . V_{\theta^T}(s_{t+1}, a_{t+1})]^2) \ (3.9)$$

The actor and critic networks have their corresponding target networks that are gradually updated after each training session, as outlined below:

$$\pi_{\theta^T} := (1-\tau).\pi_{\theta^T} + (\tau).\pi_\theta \ (3.10); \ V_{\theta^T} := (1-\tau).V_{\theta^T} + (\tau).V_\theta \ (3.11)$$

where $\pi_\theta$, $V_\theta$ are actor and critic networks respectively, and $\pi_{\theta^T}$ and $V_{\theta^T}$ are target actor and target critic networks respectively.

### 3.4.2 Hindsight Experience Replay (HER)

Hindsight Experience Replay[30] is capable of tackling sparse reward mechanisms in environments defined by Markov Decision Process. Hindsight experience replay works by replacing rewards of state action pairs sampled from the experience buffer with assigned virtual rewards. The virtual rewards are strategically sampled to suit general problems.

Andrychowicz, et al proposed Future and Final goal sampling approaches in implementing HER. In future goal sampling, the achieved goals of random states at time-steps that supersede the time-step of the current state are iteratively used in place of the desired goal of the current state. In final goal sampling however, the achieved goal of the final transition is used as the target goal for all prior transitions. In both goal sampling strategies, when the goal is changed to the sampled virtual goal, the reward value is recomputed based on the sampled goal and the achieved goal for each transition.

When using hindsight experience, it is crucial to combine the observation and the desired goal into a single state vector to be fed into the actor network, by doing so, the actor learns to effectively map correct actions capable of reaching a desired goal to any given state goal representations.

### 3.4.3 Demonstration Guided Exploration(DEX)

Demonstration guided exploration[31] is a technique that enables the Reinforcement Learning (RL) agent to learn from expert demonstrations. This technique alters the natural objective of the RL agent to ensure that it estimates actions closer to the expert actions, while also regularising the critic network to avoid overestimation of Q values.

As proposed by Huang et al.[31], DEX involves two replay buffers, with the first replay buffer containing expert demonstrations and the second one containing agent experiences. For each training iteration, a batch agent experiences and expert demonstrations are sampled. For each state-action pair $(s_t, a_t)$ corresponding to the agent experience, an expert action $a_t^{(e)}$ is computed as the weighted sum of sampled expert actions $(a^{(1)}, a^{(2)}, a^{(3)}...a^{(n)})$, whose corresponding states are $(s^{(1)}, s^{(2)}, s^{(3)}...s^{(n)})$ are closest to the sampled state $s_t$. The expert action $a^{(e)}$ can be expressed as follows:

$$a^{(e)} = \pi^{(e)}(s_t) \quad (3.12); \quad \pi^{(e)}(s_t) = \frac{\sum_{i=1}^{k} \exp(-\|s_t - s^{(i)}\|_2).a^{(i)}}{\sum_{i=1}^{k} \exp(-\|s_t - s^{(i)}\|_2)} \quad (3.13)$$

where $k$ corresponds to the number of nearest neighbours to $s_t$ from the demonstration buffer[31].

The objective of the actor network under the influence of demonstration guided exploration is similar to the objective of DDPG technique, with minor alterations as shown below:

$$\max(\mathbb{E}[V_\theta(S_t, a_t) - \alpha . d(a_t, a_t^{(e)})]) \quad (3.14) .$$

Similarly, the objective of the critic network is altered as shown below:

$$\min([V_\theta(s_t, a_t) - (r + \gamma . V_{\theta^T}(s_{t+1}, a_{t+1}) - \alpha . d(a_{t+1}, a_{t+1}^{(e)})]^2) \quad (3.15)$$

where $\alpha$ is an exploration coefficient, and $d$ is a distance function, typically the Euclidean distance. The term $\alpha . d(a_{t+1}, a_{t+1}^{(e)})$ is the regularisation term used to ensure that overestimation or inaccurate estimation of the Q values are mitigated.

## 3.5 Singular Value Decomposition

*SVD* is a *POD* technique that factorises a given $(m \times n)$ matrix into three matrices: $U$, $\Sigma$, and $V^T$ such that:

$$M = U \cdot \Sigma \cdot V^T \quad (3.16)$$

Where $U$ and $V$ are both orthogonal matrices (if $M \in R^{m \times n}$) or unitary matrices (if $M \in C^{m \times n}$) with shapes $(m \times m)$ and $(n \times n)$ respectively, and $\Sigma$ is a rectangular diagonal matrix of size $(m \times n)$ that contains the singular values. To compute $U$, we first compute the eigenvectors and the eigenvalues of the $(m \times m)$ square matrix $M \cdot M^T$. $U$ is then populated column wise in order of decreasing corresponding eigenvalues. Next, the $\Sigma$ is populated diagonally with the square root of decreasing eigenvalues. Finally, the eigenvectors of the $(n \times n)$ square matrix $M^T \cdot M$ are computed and used to create $V$. One aspect to note is that $M^T \cdot M$ and $M \cdot M^T$ have the same eigenvalues but different eigenvectors. In our study, we computed the r-rank decomposition of model weights, such that for a given size of the weight $(m \times n)$, we approximate it as:

$$w = U^{m \times r} \cdot \Sigma^{r \times r} \cdot V^{T^{r \times n}}; \quad where \ \ r \leq min(m, n) \quad (3.17)$$

When conducting low-rank approximation experiments, it is important to determine appropriate range of the ranks for singular value decomposition, as they can vary based on the characteristics of any given environment.

# 4  $L_0$ Norm Regularization for Policy Sparsification

## 4.1  $L_0$ Norm Regularization - Background

The base framework for $L_0$ norm regularization to achieve sparsification was provided by Louizos et al [32]. We also studied similar frameworks suggested by Srinivas et al.[33] and leveraged common aspects from both of these studies. As suggested by Louizos et al.[32], the goal is to estimate an optimal sparsity map $z$ (referred to as gate by Louizos et al.[32]) with lesser number of non-zero elements, which can transform the dense policy, during model training, to a sparse one via a hadamard product. The following equations for $z$ and $s$ are given by Louizos et al.[32], where $z$ is a hard sigmoid rectification of $s$:

$$z = \min(1, \max(0, \bar{s})) \qquad (4.1)$$

where $s$ is a continuous random variable in the interval (0,1) with a probability density of $q_s(s|\phi)$ and cumulative density of $Q_s(s|\phi)$. The parameters of the distribution $\phi$ are given as $\log(\alpha)$ denoting the location and $\beta$ denoting the temperature. The distribution is further stretched to the interval $(\gamma, \zeta)$ where $\gamma < 0$ and $\zeta > 1$ and then a hard-sigmoid is applied on its random samples as shown below:

$$u \sim U(0,1), \bar{s} = \sigma\left(\frac{\log(u) - \log(1-u) + \log(\alpha)}{\beta}\right) \cdot (\zeta - \gamma) + \gamma \qquad (4.2)$$

At inference, the Louizos et al. proposed that $\hat{z}$ be used instead of $z$. The expression for $\hat{z}$ is given as:

$$\hat{z} = \min(1, \max(0, \sigma(\log \alpha) \cdot (\zeta\text{-}\gamma) + \gamma)) \qquad (4.3)$$

where: $\sigma$ represents a sigmoid operation.

Using the above, Louizos et al.[32] proposed to add a sparsity penalty term $L_{sp}$, governed by a penalty coefficient $\lambda_c$ to be minimised alongside the network's cost function to induce sparsity. The coefficient $\lambda_c$, a weighting factor, aids in achieving optimum sparsification. For a parameter vector with dimensionality $|w|$, the expression for $L_{sp}$[32] is:

$$L_{sp} = \sum_{j=1}^{|w|} \sigma(log\alpha_j - \beta log \frac{-\gamma}{\zeta}) \qquad (4.4)$$

## 4.2 Custom $L_0$ Norm Regularization - Key Differentiators

Previous studies involving regularization techniques targeted sparse auto-encoders[34] and other deep neural networks for image classification problems[32]. However, based on our knowledge, this is the first-of-its-kind implementation of $L_0$ norm regularization technique with singular value decomposition to a lower rank, to achieve greater sparsity and added compression, without decay in rewards, for deep reinforcement learning policies in several complex environments. Additionally, Our technical implementation of the $L_0$ norm regularization technique differs from the previous studies where we retained all of the contributions to the sparsity term to be the same during training and evaluation. During experimentation, we found that retaining all contributions during evaluation induced more sparsity in the reinforcement learning policies. Consequently, we did not use a separate equation $\hat{z}$ for evaluation. Once the sparsity map is determined, the sparse weights $\hat{w}$ are calculated with an element-wise product of the computed optimal sparsity map $z$ and the dense weight matrix as shown below:

$$\hat{w} = z \cdot w \qquad (5.1)$$

. Additionally, we custom developed the following deep reinforcement learning algorithms for dense and sparse policies for implementation in multiple environments - which is detailed in subsequent sections.

## 4.3 *DQN / DDQN* Algorithm with Sparsification

Let $\pi_\theta$ be the action logits of the policy, $L_{\text{SP}}$ the sparsification penalty, $\lambda_c$ the sparsification coefficient, $b$ the batch size or number of samples, and $\epsilon$ the exploration rate.

---
**Algorithm 1** DQN and DDQN Algorithm with Sparsification
---
 1: Initialize $\epsilon$, $\epsilon_{\max}$, $\epsilon_{\min}$
 2: Initialize policy and target networks with same parameters
 3: Initialize replay buffer
 4: **for** episode $= 1, 2, 3, \ldots$ **do**
 5:     **if** $\epsilon \to \epsilon_{\max}$ **then**
 6:         Explore environment more with policy
 7:     **else if** $\epsilon \to \epsilon_{\min}$ **then**
 8:         Exploit environment more with policy
 9:     **end if**
10:     Gather experience $(s_t, a_t, r_t, s_{t+1})$ and store in replay buffer
11:     **if** replay buffer size $> b$ **then**
12:         Sample random batch of experience from replay buffer
13:         **if** sparsity_flag is True **then**
14:             Compute $\pi_\theta$ and $L_{\text{SP}}$ from the batch of sampled $s_t$
15:         **else**
16:             Compute $\pi_\theta$ only from the batch of sampled $s_t$
17:         **end if**
18:         Gather batch of $q_t$ corresponding to batch $a_t$
19:         Compute $q_{\text{future}}$ from batch of sampled $s_{t+1}$
20:         Compute $q_{\text{optimal}}$ with Bellman's equation
21:         Compute $L_\theta = (q_t - q_{\text{optimal}})^2$
22:         **if** sparsity_flag is True **then**
23:             Compute $L_\theta := L_\theta + \left(\frac{\lambda_c \cdot L_{\text{SP}}}{b}\right)$
24:         **end if**
25:         Minimise $L_\theta$
26:         Update policy
27:     **end if**
28:     Decay $\epsilon$
29: **end for**
---

## 4.4 *PPO* Algorithm with Sparsification

Let $\pi_\theta$ be the action logits of policy, $L_{\text{SP}}$ the sparsification penalty, $\lambda_c$ the sparsification coefficient, $b$ the batch size or number of samples, $N$ the number of rollouts, $P$ the number of training epochs for policy, $H$ the Entropy, $\lambda_h$ the entropy coefficient and $\epsilon = 0.2$.

---
**Algorithm 2** PPO Algorithm with Sparsification
---
1: Initialize policy network $\pi_\theta$ with weights $\theta$
2: **for** episode $= 0, 1, 2, \ldots$ **do**
3:     **for** $t = 0, 1, 2, \ldots, N-1$ **do**
4:         Run policy $\pi_\theta$ and value networks in environment for $T$ timesteps
5:         Gather experience $(s_t, a_t, r_t, s_{t+1})$
6:         Compute advantage estimates $\hat{A}_t$
7:     **end for**
8:     **for** $p = 0, 1, 2, \ldots, P-1$ **do**
9:         **if** sparsity_flag is True **then**
10:             Compute $\pi_\theta$ and $L_{\text{SP}}$ from the gathered $s_t$
11:         **else**
12:             Compute $\pi_\theta$ only from the gathered $s_t$
13:         **end if**
14:         Compute $L^{\text{CPI}} = r_t(\theta) \cdot \hat{A}_t$
15:         Compute $L^{\text{clip}} = \text{clip}(r_t(\theta), 1-\epsilon, 1+\epsilon) \cdot \hat{A}_t$
16:         Compute $L_\theta = -\min(L^{\text{CPI}}, L^{\text{clip}})$
17:         **if** sparsity_flag is True **then**
18:             Compute $L_\theta := L_\theta + \left(\frac{\lambda_c \cdot L_{\text{SP}}}{b}\right)$
19:         **end if**
20:         Compute $L_\theta = L_\theta + \lambda_h \cdot H(\pi_\theta)$
21:         Minimise $L_\theta$
22:         Update policy by applying gradient descent to $\theta$
23:     **end for**
24: **end for**

**Algorithm 3** DDPG Algorithm With Sparsity, HER, and DEX
1: Initialize environment, agent replay buffer $D_A$, and expert replay buffer $D_E$
2: Initialize actor network $\pi_\theta$, critic network $V_{theta}$, target actor network $\pi_{\theta T} = \pi_\theta$, target critic network $V_{\theta T} = \theta$, max # of episodes $N$, batch size b, and counter $= 0$
3: repeat
    a. counter:= counter+1
    b. for t in $1, 2, 3..., T$
        i. Observe state $s_t$, desired goal $g$, achieved goal $g'_{t+1}$ and select action $a_t = \pi_\theta(s_t||g'_t) + \epsilon \cdot (0.1 \cdot N(0,1))$
        ii. Execute action $a_t$ on environment
        iii. Observe reward $r$, next state $s_{t+1}$, achieved goal $g'_{t+1}$ and $d$ (terminal state)
        iv. Register experience $e = (s_t||g'_t, a_t, r, s_{t+1}||g_{t+1}\prime, g, d)$ in $D_A$
        v. if size of $D_A \geq b$
            1. HER Sample batch of agent experience $e_b \sim D_A$
            2. Sample expert demonstrations $e'_b \sim D_E$
            3. Formulate expert policy $\pi^{(e)}$ for each $s||g$ in $e_b$ and $(s_i, a_i \in e'_b)$
            4. Initialize $s_{p|\pi_\theta}$ and $s_{p|V_\theta}$
            5. If $\pi_\theta$ is sparse
                a. $a'_n, s_{p|\pi_\theta} = \pi_\theta(s_n||g); a'_{n+1} = \pi_{\theta T}(s_{n+1}||g)$
            6. else
                a. $a'_n = \pi_\theta(s_n||g); a'_{n+1} = \pi_{\theta T}(s_{n+1}||g)$
            7. end if
            8. if $V_\theta$ is sparse
                a. $Q_n, s_{p|V_\theta} = V_\theta(s_n||g, a_n)$
                b. $Q_{n+1}, s_{p|V_{\theta T}} = V_\theta(s_{n+1}||g, a_{n+1}\prime)$
            9. else
                a. $Q_n = V_\theta(s_n||g, a_n)$
                b. $Q_{n+1} = V_{\theta T}(s_{n+1}||g, a_{n+1}\prime)$
            10. end if
            11. Minimize:

$$-\nabla_\theta \frac{1}{|b|} \sum_{s_n||g \in \epsilon_b} V_\theta(s_n||g, a'_n) - \alpha d(a'_n, \pi^{(e)}(s_n||g)) + \frac{1}{b}\lambda_s \cdot s_{p|\pi_\theta}$$

            12. Minimize:

$$\nabla \frac{1}{|b|} \sum_{s_{n+1}||g \in \epsilon_b} |Q_n - (r + \gamma Q_{n+1} - \alpha d(a_{n+1}, \pi^{(e)}(s_{n+1}||g)))|^2 + \frac{1}{|b|}\lambda_s \cdot s_{p|V_\theta}$$

        vi. end if
    c. end for
12. until counter $= N$

# 5 Experimental Results

## 5.1 Policy Sparsification

### 5.1.1 Initial Benchmarking Studies: Cartpole, Acrobot and LunarLander

In our first set of benchmarking experiments, we implemented our custom $L_0$ norm regularization technique in three different OpenAI's open source gym environments with multi-dimensional continuous observation spaces and discrete action spaces with varying degrees of complexity: Cartpole-v1; Acrobot-v1; and LunarLander-v2. For each of these environments, we developed: dense; sparse; decomposed dense; and decomposed sparse policies using two different reinforcement learning frameworks: *DQN* and *PPO*.

In these benchmarking studies, we experimented with 10 different sparsification coefficients which yielded 10 different sparse policies for each combination of the environment (Cartpole vs Acrobot vs LunarLander) and the reinforcement learning framework (*DQN* vs *PPO*). We evaluated the sparsity and rewards and retained the best performing sparse policy which maximized the objective $r + S_p$, where $r$ is the reward and $S_p$ is the percentage sparsity of the policy for further analysis and low rank decomposition. The objective is to consider such a sparse policy, from the entire sample of 10 sparse policies that had maximum sparsity with optimum rewards for low rank decomposition. We set target rewards for each environment where policy training terminates upon reaching or crossing this target.

| Environment | Target Reward |
|---|---|
| Cartpole-v1 | 500 |
| Acrobot-v1 | -100 |
| LunarLander-v2 | 200 |

Table 1: Environments and Target Rewards

The training process for each of the environments with different sparsity coefficients and their corresponding performance along with the heat-maps indicating the impact of the sparsity coefficients on various performance measures is outlined in the Appendix A.1. The key performance metrics of dense and best performing sparse policies across the three environments are illustrated in Table 2, and the corresponding reward plots are outlined in the following Figures (1-6).

| Cartpole-v1 Environment | | | | | |
|---|---|---|---|---|---|
| Policy | Sparsity | Coeff ($\lambda_{sp}$) | Eval. Reward | Conv. Steps | Train. Steps |
| Dense DQN | — | — | 500.0 | 1116 | 2000 |
| Sparse DQN | 28.45 | $5.0e^{-5}$ | 500.0 | 1511 | 2000 |
| Dense PPO | — | — | 500.0 | 27 | 400 |
| Sparse PPO | 24.06 | $5.0e^{-1}$ | 500.0 | 95 | 400 |
| Acrobot-v1 Environment | | | | | |
| Policy | Sparsity | Coeff ($\lambda_{sp}$) | Eval. Reward | Conv. Steps | Train. Steps |
| Dense DQN | — | — | -99.0 | 998 | 2000 |
| Sparse DQN | 45.77 | $5.0e^{-3}$ | -95.0 | 1037 | 1000 |
| Dense PPO | — | — | -127.0 | 53 | 2000 |
| Sparse PPO | 24.37 | $5.0e^{-3}$ | -91.0 | 187 | 1000 |
| LunarLander-v2 Environment | | | | | |
| Policy | Sparsity | Coeff ($\lambda_{sp}$) | Eval. Reward | Conv. Steps | Train. Steps |
| Dense DQN | — | — | 261.3 | 2446 | 4000 |
| Sparse DQN | 47.57 | $5.0e^{-3}$ | 262.6 | 2687 | 4000 |
| Dense PPO | — | — | 201.3 | 1397 | 4000 |
| Sparse PPO | 52.65 | $1.0e^{-3}$ | 205.5 | 3640 | 4000 |

Table 2: Performance Comparison - Dense vs. Sparse Policies

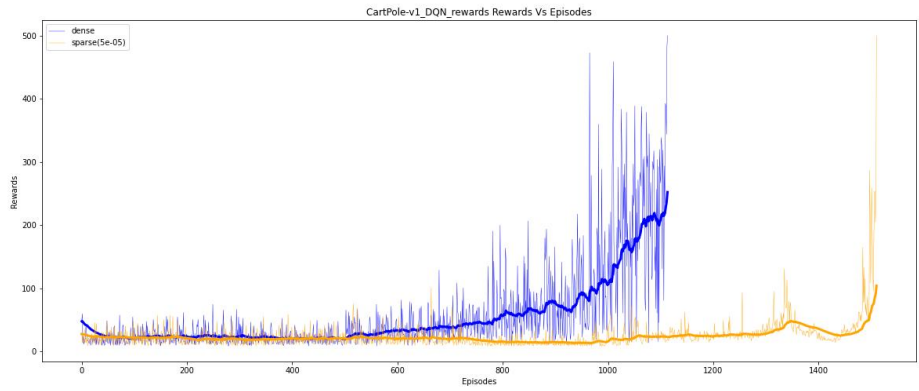

Figure 1: Reward Plots for the DQN policies in the CartPole-v1 environment.

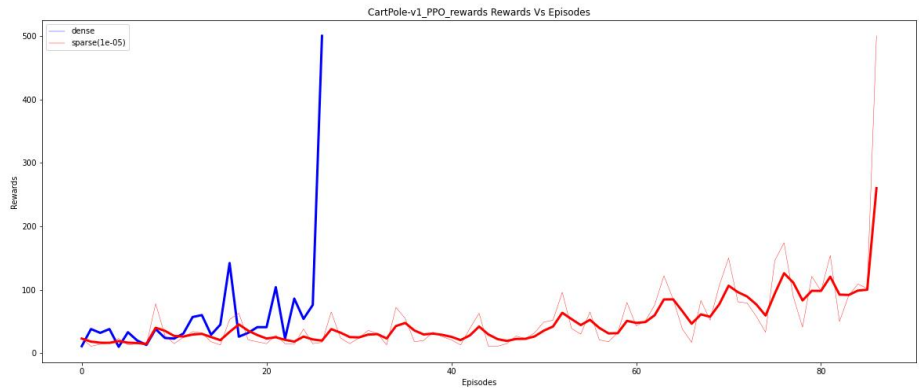

Figure 2: Reward Plots for the PPO policies in the CartPole-v1 environment.

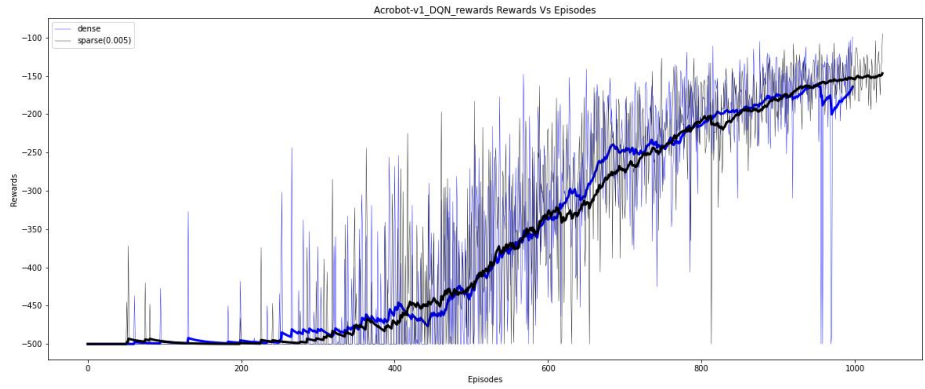

Figure 3: Reward Plots for the DQN policies in the Acrobot-v1 environment.

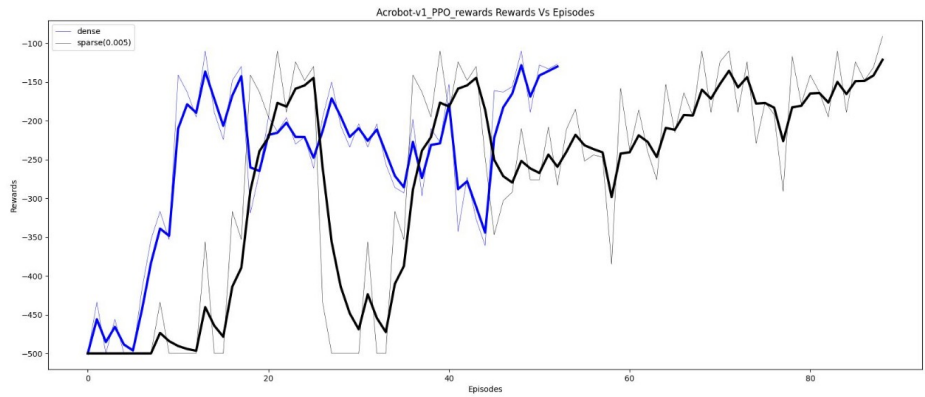

Figure 4: Reward Plots for the PPO policies in the Acrobot-v1 environment.

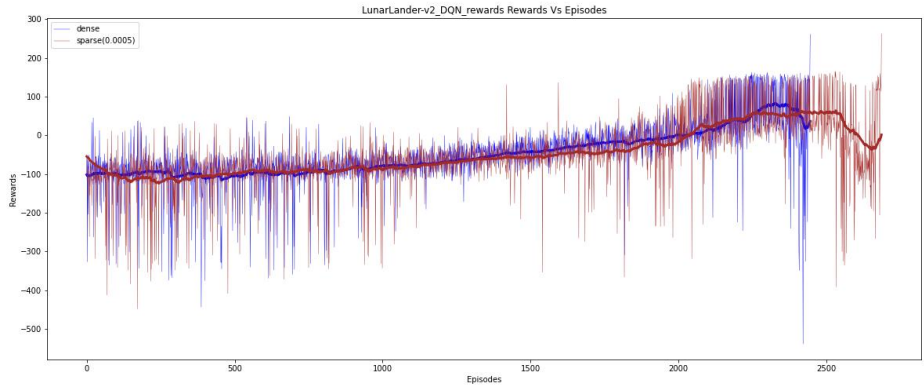

Figure 5: Reward Plots for the DQN policies in the LunarLander-v2 environment.

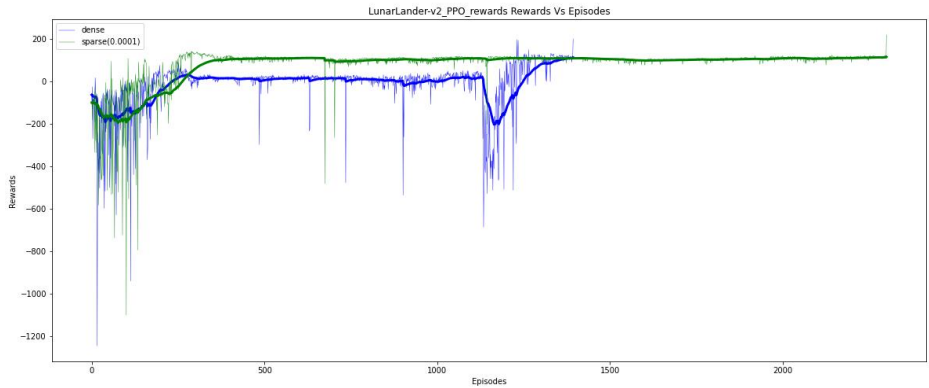

Figure 6: Reward Plots for the PPO policies in the LunarLander-v2 environment.

From the above information, we observed that the sparse policies not only retained relative performance but also exceeded the rewards in some cases, when compared to those of their dense counterparts. For e.g. the sparse DQN and PPO policies for Cartpole-v1 environment achieved the same reward of 500 when compared to the corresponding dense policies whereas the sparse DQN and PPO policies for the Acrobot-v1 environment outperformed their dense counterparts by achieving higher rewards of -95 vs -99 and -91 vs -127 respectively. We observed the same phenomenon in the moderately complex environment of LunarLander-v2 where the DQN and PPO sparse policies consistently outperformed the corresponding dense policies. These results provide solid evidence that by obtaining optimal sparsity one can not only engineer leaner deep reinforcement learning policies but also gain improvement in rewards.

We also observed that sparse policies, in some environments with simpler action spaces, required greater number of of steps to converge when compared to their dense counterparts, although the convergence was well within the range of the training steps. In some environments the sparse policies converged either at a similar rate or faster (e.g. LunarLander PPO) when compared to their dense counterparts and the performance of sparse policies was as good or better for equal number of training steps. Based on theoretical and mathematical characteristics we attribute the greater number of steps for sparse policy convergence to the nature of the $L_0$ norm regularization using the layer based technique. We however, did not find any correlation or credible reasoning for faster convergence of sparse policy in certain environments. We can further visualize the effects of policy sparsification on various measures of various policies in the Heatmap figures provided in the Appendix A2.

### 5.1.2 SuperMarioBros - A Complex Environment

Based on the successful implementation of $L_0$ norm regularization technique to obtain optimal sparsity in deep reinforcement learning polices in moderately complex environments we extended our experiments to a more complex environment to further bolster our claim that optimally sparse policies obtained via $L_0$ norm regularization will perform as good or better than their dense counterparts while yielding equivalent or greater rewards when compared to their dense counterparts.

For this experiment we used the "SuperMarioBros 7.1.0[35]" game environment, with the action space reduced to 5 ("RIGHT ONLY"). In this environment, each observation is a stack of 4 gray scale frames with pixel values normalised to a range between 0 and 1. A given action was repeated on 4 observations and their rewards were accumulated and averaged across those observations and the corresponding next observation (the maximum of pixel values of the last two observations). Each policy implementation was trained for 20,000 episodes, and each episode spanned a maximum of 500 steps. During training, policy evaluation was carried out after every 10 episodes. We conducted policy sparsification experiments with the DDQN and PPO reinforcement learning algorithms on the SuperMarioBros 7.1.0 environment. We chose the DDQN algorithm instead of the DQN algorithm for this environment to reduce the overestimation bias. The following Figures 7-9 illustrate the performance of the dense DDQN and sparse DDQN policies (with different sparsity coefficients) in the SuperMarioBros 7.1.0 environment. The results of the experiments with the PPO algorithm in the SuperMarioBros 7.1.0 environment are outlined in Appendix A3.

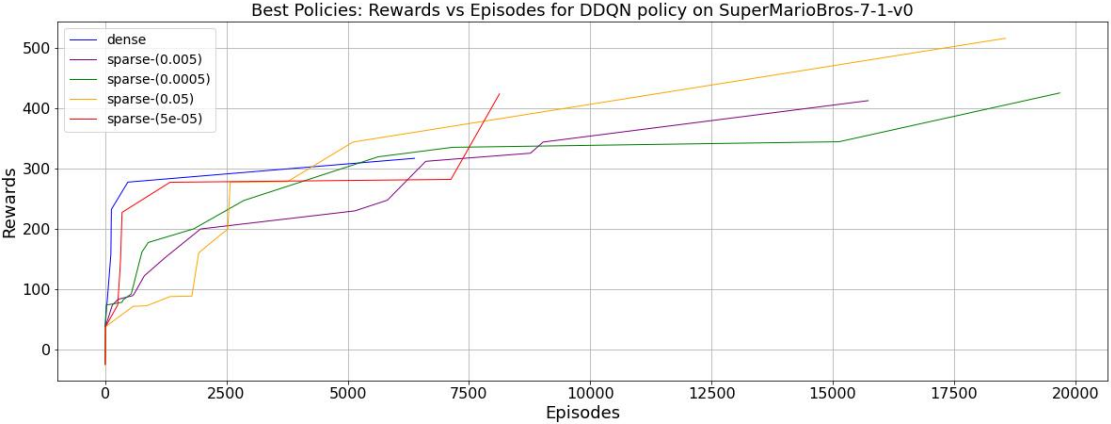

Figure 7: Episodic Rewards for DDQN Dense and Sparse policies - SuperMarioBros 7.1.0.

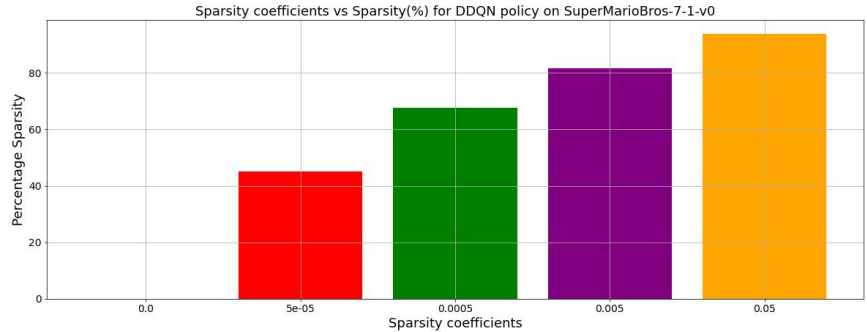

Figure 8: Sparsity induced by various sparsification coefficients for DDQN policies.

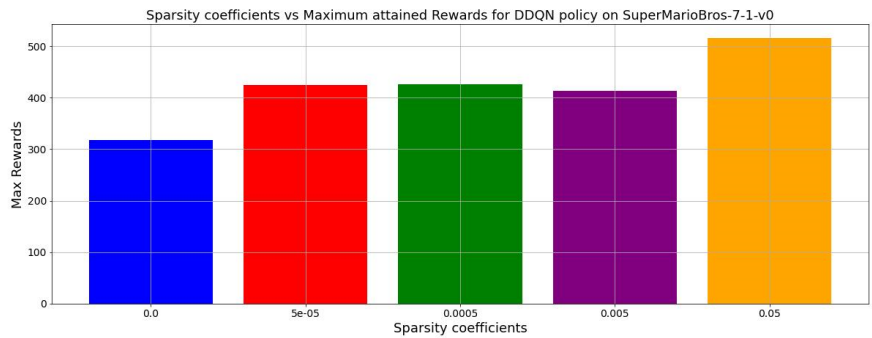

Figure 9: Max rewards for the dense and sparse DDQN policies.

We observed in the DDQN implementation with priority experience that the sparse policies outperformed the dense policy consistently and by a greater degree. Additionally, the best performing sparse policy was also the one with the greatest sparsity induced. The average trend-lines for all policies tended upwards, indicating that the policies were converging and that the performance could be improved by extending the training for more episodes. This experiment in a more complex SuperMarioBros 7.1.0 environment provides convincing evidence that optimally sparse policies, obtained via $L_0$ regularization, can perform as good or better than the dense policies while yielding inherent benefits of limiting overfitting and gaining computing efficiencies, especially in more complex environments.

### 5.1.3 Surgical Robot Learning (SurROL) - Complex Environment With Continuous Action Space

In addition to conducting experiments in environments with discrete action space, we extended the implementation of our $L_0$ norm regularization technique to a promising robotic surgery application with a continuous action space. SurROL is a very complex Reinforcement Learning Centered dVRK (Da Vinci Research Kit) compatible platform for Surgical Robot Learning (SurROL)[36], with continuous action space, that supports more realistic physical interactions including Patient Side Manipulators (PSM) and Endoscopic Camera Manipulator (ECM) scenarios.

SurROL features several tasks covering different levels of surgical skills. We conducted our experiments in the NeedleReach, which is a continuous control and a goal based environment. Due to the continuous control nature of the environment, we leveraged an off-policy reinforcement learning technique - Deep Deterministic Policy Gradient (DDPG) along with Hindsight Experience Replay (HER) and Demonstration Guided Exploration (DEX) to achieve faster convergence.

For this experiment, the actor neural network comprises 4 fully connected layers. The first layer has an output size of 256, the two hidden layers also output 256 neurons and the final layer outputs the size of the continuous action space. The critic network features similar architecture, where the input layer is fed the state goal representation and the corresponding estimated action to estimate the corresponding Q value. In our implementation, we used a batch size of 64 during training and an experience buffer size of 10,000 for the agent, and 5,000 for the expert demonstration. Both the actor and critic newtorks use a learning rate of $10^{-5}$. To promote exploration, we added a scaled random gaussian noise to every estimated action, with a scale factor of 0.1. The noise was further scaled by an epsilon value that decayed linearly from 1.0 to 0.0 throughout the course of training. We selected the exploration coefficient $\alpha$ to be 5 and the number of nearest neighbours $k$ to be 5 as well, as suggested by Huang et al[31]. To update the target networks, we selected $\tau$ to be $10^{-3}$. We used the future goal sampling strategy for the hindsight experience replay, with $k_{future} = 4$.

We trained the dense policy and several sparse policies for the course of 200 episodes (the policy is updated 50 times per episode). We used sparsification coefficients of $5.0e^{-05}$, $5.0e^{-04}$, $5.0e^{-03}$, $5.0e^{-02}$, $5.0e^{-01}$, $1.0e^{-00}$, $2.0e^{-00}$ and $5.0e^{-00}$.

As shown below in the following figures, the $L_0$ norm regularization technique achieved achieving greater sparsity without decay in rewards.

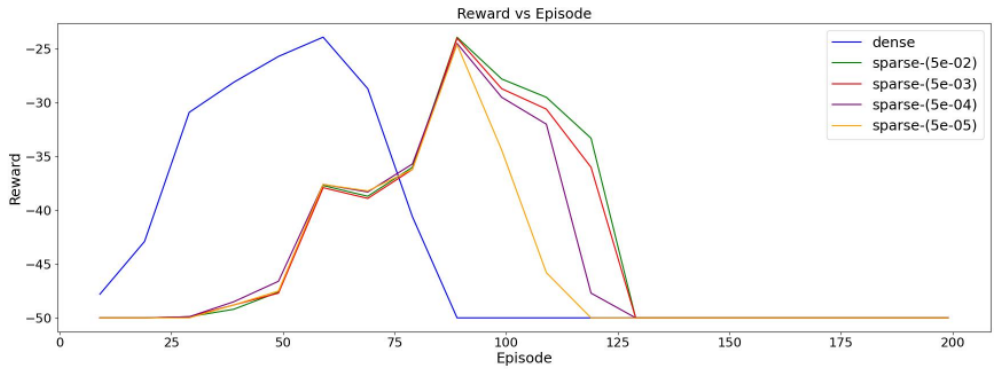

Figure 10: Max rewards for the dense and sparse DDQN policies.

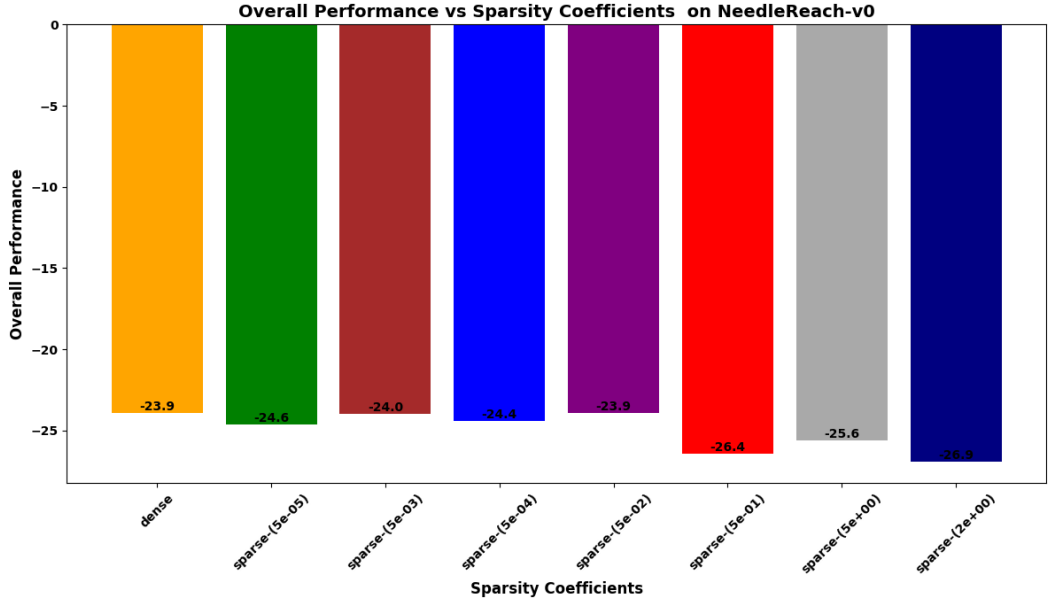

Figure 11: Overall Performance by Sparsity Coefficients.

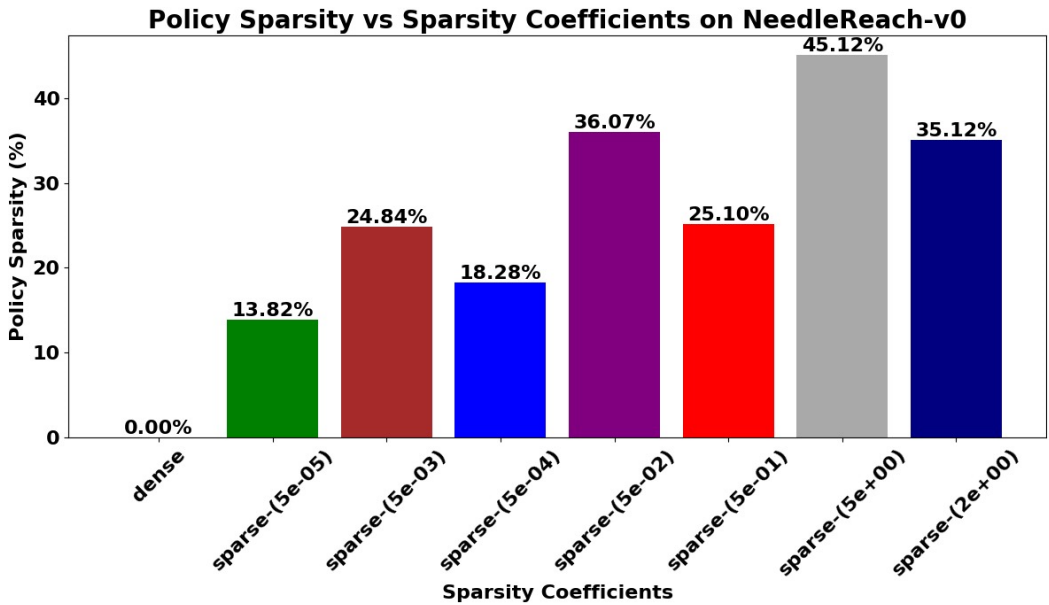

Figure 12: Percentage of Sparsity by Sparsity Coefficients.

From the figures above, it is evident that the sparse policy with the sparsity coefficient of $5.0e^{-02}$ retained the performance (rewards) while achieving 36% policy sparsification in less than half of the total time-steps, which is significant for a continuous action space environ-ments. Moreover, the relative performance of several other sparse policies was comparable to that of the dense policy. We also observed that irrespective of the magnitudinal differences amongst the sparsity coefficients, the relative performance did not change significantly. We attribute this behavior towards the implementation of the demonstrated guided exploration which tends to prioritize expected return over the sparsification penalty, as well as the lower learning rate of $10.0e^{-05}$, which was chosen to facilitate policy convergence, as the policy could not converge with faster learning rates. The results of this experiment provides further evidence that the optimally sparse policies obtained via $L_0$ regularization can achieve computational efficiencies while performing as good or better than the dense policies.

## 5.2 Comparison of $L_0$, $L_1$ and $L_2$ Regularization

Beyond implementing the $L_0$ norm regularization technique, we experimented with $L_1$ and $L_2$ regularization techniques to induce sparsity and compared their relative performance in the compex SuperMarioBros 7.1.0 environment. Based on the results of our experiments, we found that $L_0$ norm regularization not only induced highest sparsity but also delivered greatest rewards, as illustrated in the following figures.

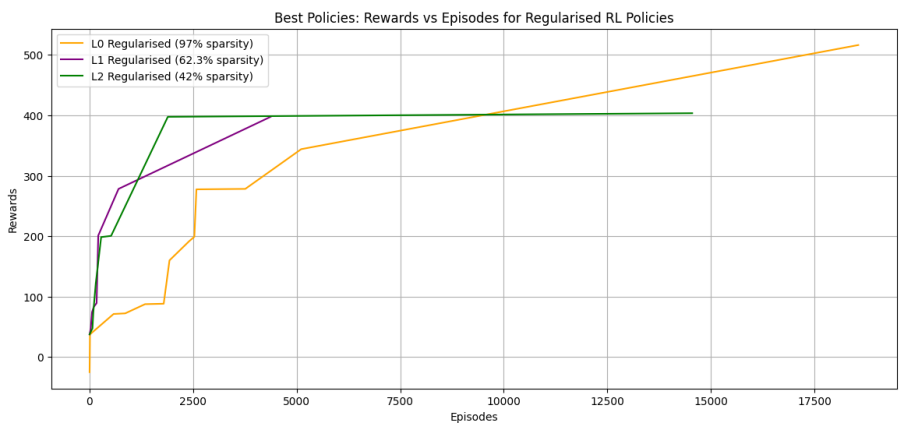

Figure 13: Episodic Rewards - $L_0$ vs $L_1$ vs $L_2$ regularization for the DDQN Policy.

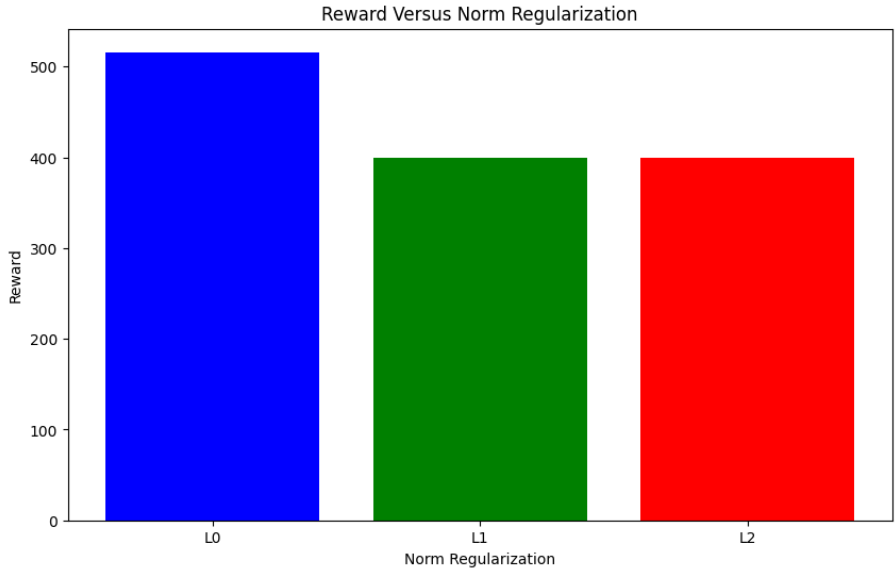

Figure 14: Max Rewards - $L_0$ vs $L_1$ vs $L_2$ regularization for the DDQN Policy.

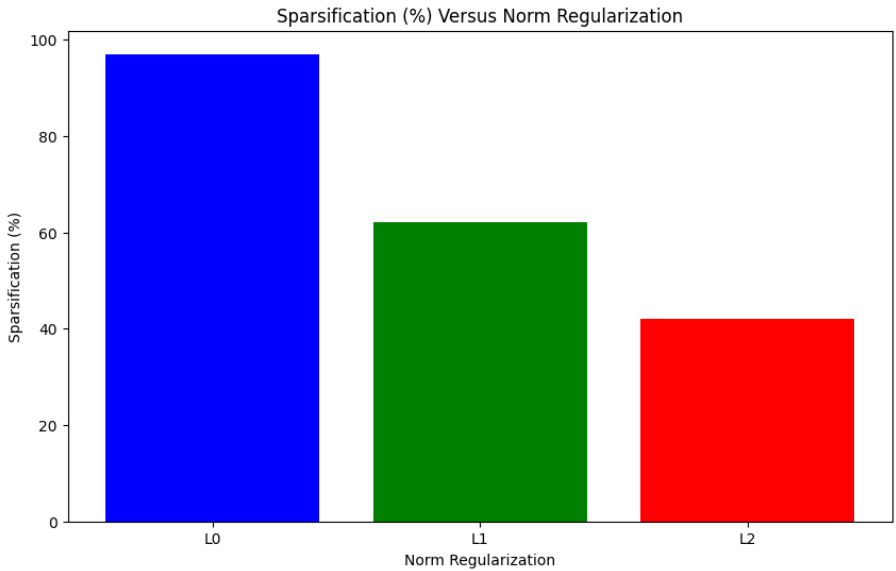

Figure 15: Max Sparsity - $L_0$ vs $L_1$ vs $L_2$ regularization for the DDQN Policy.

## 5.3 Low Rank Decomposition

The secondary goal of our experiments is to analyze the impact of sparsity on low-rank decomposition. We tested the performance of the dense and the best performing sparse policies, for different ranks: $r \in U\{5, 6, ..., 28, 29\}$ for Cartpole-v1, Acrobot-v1 and LunarLander-v2 environments; and $r \in U\{10, 11, ..., 180, 190\}$ for SuperMarioBros 7.1.0 - due to its' characteristics with convolutional layers. We analyzed the effects of policy decomposition on various measures (number of trainable parameters and rewards) of the dense and sparse policies across the above mentioned environments. It is worth noting that the policies used in this implementation had similar architecture which consisted of an input layer with same input neurons as the observation space, a hidden layer of 128 neurons, and an output layer with neurons equivalent to the action space. Despite having the similar architecture, the size of each policy is dependent on the environment, since the environments had different observation and action spaces. The following Table 3 outlines the observation space, action space, and policy sizes of various environments used in the experiments.

| Environment | Observation Space | Action Space | Parameters |
|---|---|---|---|
| Cartpole-v1 | 4 | 2 | 11,042 |
| Acrobot-v1 | 6 | 3 | 11,331 |
| LunarLander-v2 | 8 | 4 | 11,620 |
| SuperMarioBros 7.1.0 | $4 \times 84 \times 84$ (pixel space) | 5 | 4,015,093 |

Table 3: Environment Space and Policy Size

We illustrated the results of the lowest ranks of the decomposed (compressed) policies that crossed the reward thresholds of several environments in the following Table 4.

The compression of policies for different ranks for several environments is detailed in Appendix A4. We did not exceed a rank value of 29 for Cartpole-v1, Acrobot-v1 and LunarLander-v2 environments as such a rank would have more matrix elements than the dense policy, hence defeating the purpose of the low rank decomposition and approximation. Similarly, we did not exceed a rank value of 190 for the SuperMarioBros 7.1.0 environment.

It is well known that in the context of low rank approximation, the higher the rank, the better the approximation. However it is worth noting that we observed that some of the low rank approximated policies outperforming approximated policies of higher ranks, especially in the Acrobot-v1 and LunarLander-v2 environments.

| Cartpole-v1 Environment | | | |
|---|---|---|---|
| Policy | Rank | Size Decrease (%) | Reward |
| Dense DQN | 8 | 70.35 | 500.0 |
| Sparse DQN | 7 | 73.23 | 500.0 |
| Dense PPO | 5 | 78.88 | 500.0 |
| Sparse PPO | 5 | 78.88 | 500.0 |
| Acrobot-v1 Environment | | | |
| Policy | Rank | Size Decrease (%) | Reward |
| Dense DQN | 27 | 8.08 | -85.0 |
| Sparse DQN | 15 | 47.48 | -82.0 |
| Dense PPO | 16 | 44.39 | -73.0 |
| Sparse PPO | 13 | 53.55 | -73.0 |
| LunarLander-v2 Environment | | | |
| Policy | Rank | Size Decrease (%) | Reward |
| Dense DQN | 22 | 23.89 | 260.89 |
| Sparse DQN | 22 | 23.89 | 279.01 |
| Dense PPO | 10 | 60.24 | 207.41 |
| Sparse PPO | 6 | 73.84 | 209.15 |
| SuperMarioBros 7.1.0 Environment | | | |
| Policy | Rank | Size Decrease (%) | Reward |
| Dense DDQN | 170 | 29.02 | 317.0 |
| Sparse DDQN | 70 | 70.07 | 516.25 |

Table 4: Performance of low-rank approximated dense and sparse policies

The impact of sparsity on low-rank decomposition was even more evident in the complex environments with both the discrete action space in SuperMarioBros environment and the continuous action space in the Surgical Robot Learning(SurROL) environment. In the SuperMarioBros environment, the performance of sparse policies was noteworthy where the decomposed sparse policy achieved 70% compression while outperforming the decomposed dense policy by 200 points. In the Surgical Robot Learning environment with a continuous action space, none of the dense policies were performant, when subjected to low-rank decomposition. Only the sparse policies were performant after low-rank decomposition, illustrat-ing the effectiveness of sparsity in gaining computational efficiencies and model compression without decay in rewards. The following figures outline the performance of the decomposed dense policy by rank and the decomposed sparse policy (the best performing sparse policy with sparsity coefficient of $5.0e^{-02}$).

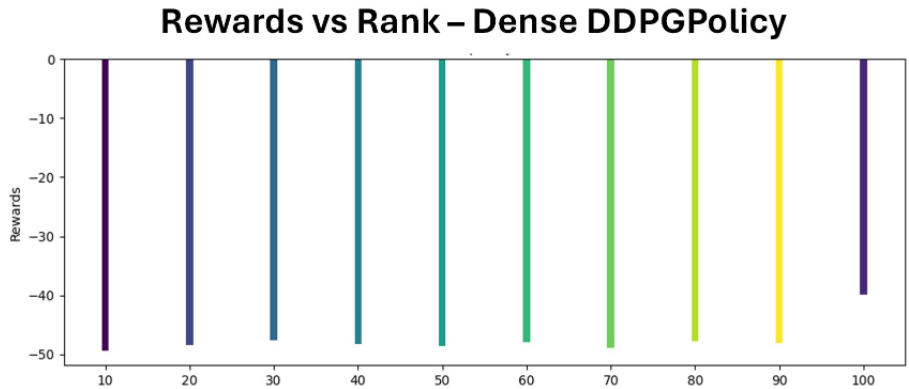

Figure 16: Rewards by Rank - Dense DDPG policy in SurROL NeedleReach environment.

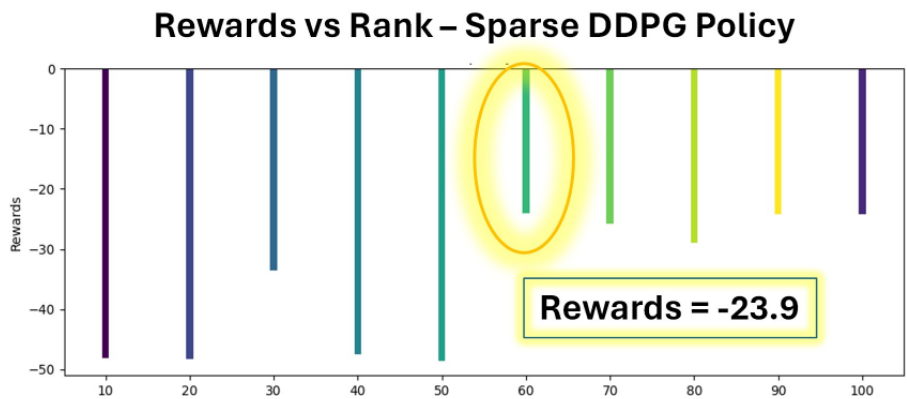

Figure 17: Rewards by Rank - Sparse DDPG policy in SurROL NeedleReach environment.

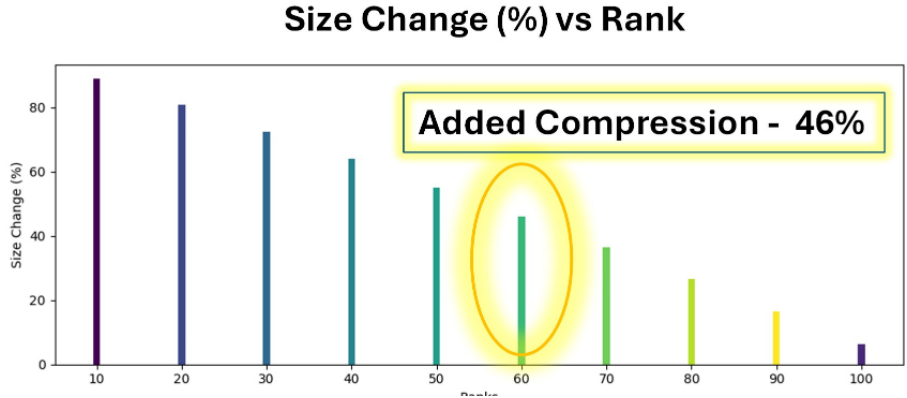

Figure 18: Percentage of Size Change by Rank - Sparse DDPG policy in SurROL NeedleReach environment.

# 6 Conclusion

$L_0$ norm regularization is an effective means to achieve optimal sparsification of deep reinforcement learning policies to limit overfitting and reduce computational cost during inference. From our experimental results we observed that several sparse policies, even with sparsification above 50% of the policy parameters, were able to attain maximum rewards and converge appropriately when compared to their dense counterparts. We also note that the number of iterations required for convergence are some times higher for sparse policies. Moreover, sparse policies promote decomposition of neural network based reinforcement learning policies to a lower rank and hence achieve greater compression when compared to the compression achieved by dense policies. This is very beneficial when dealing with edge computing applications, especially in healthcare. It is important to note that the best rank for the low rank approximation needs to be decided based on experimentation - to evaluate the trade off between policy size and performance.

As for future works; we hope to devise and explore better and more efficient sparsification techniques, such as the $L_0$ norm group sparsification, which may potentially lead to a more computationally efficient technique than the layer based $L_0$ norm sparsification.

Additionally, we hope to explore combining different regularization methods to gain optimum sparsity and improved performance especially when dealing with complex and high dimensional environments.

# 7 Glossary

1. *DQN*: Deep Q Network

2. *DDQN*: Double Deep Q Network

3. *PPO*: Proximal Policy Optimization

4. *POD*: Proper Orthogonal Decomposition

5. *SVD*: Singular Value Decomposition

6. *GAE*: Generalised Advantage Estimate

7. *KLD*: Kullback Leibler Divergence

8. *CPI*: Conservative Policy Iteration

9. *MDP*: Markov Decision Process

10. *DDPG*: Deep Deterministic Policy Gradient

11. *HER*: Hindsight Experience Replay

12. *DEX*: Demonstration Guided Exploration

# A Appendix

## A.1 Sparse Policy Rewards for Different Sparse Coefficients

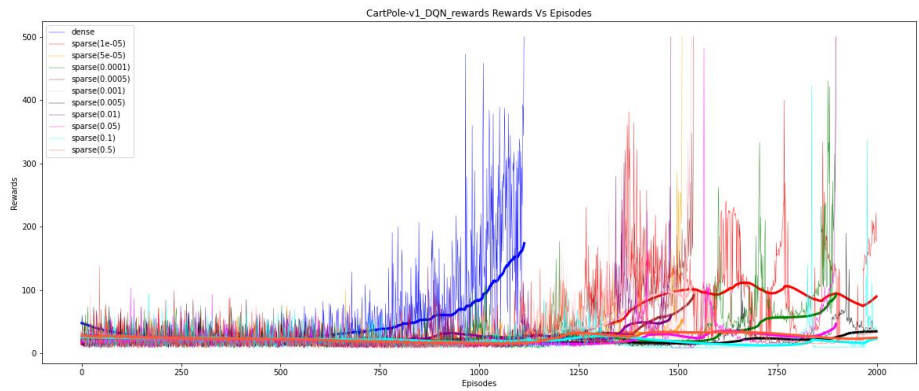

Figure 19: Reward Plots for the DQN policies in the CartPole-v1 environment.

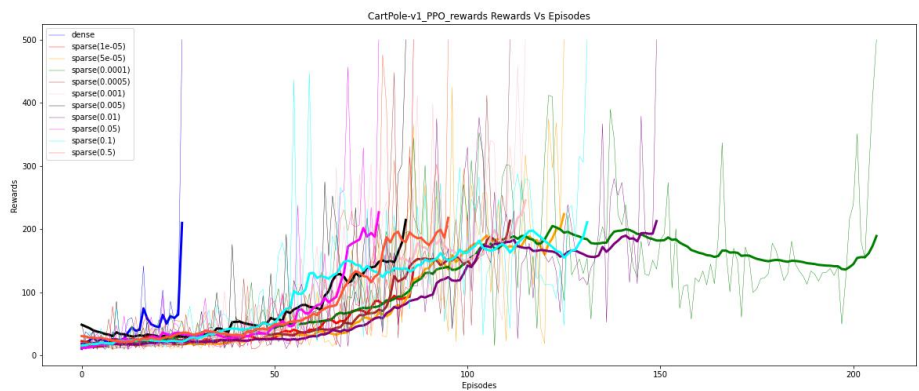

Figure 20: Reward Plots for the PPO policies in the CartPole-v1 environment.

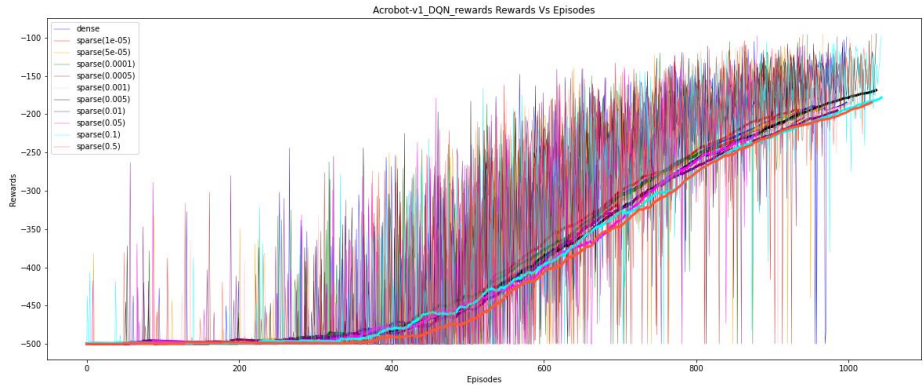

Figure 21: Reward Plots for the DQN policies in the Acrobot-v1 environment.

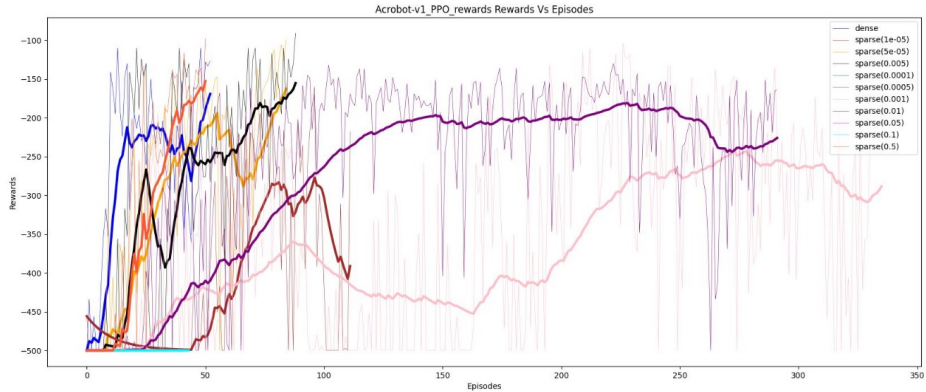

Figure 22: Reward Plots for the PPO policies in the Acrobot-v1 environment.

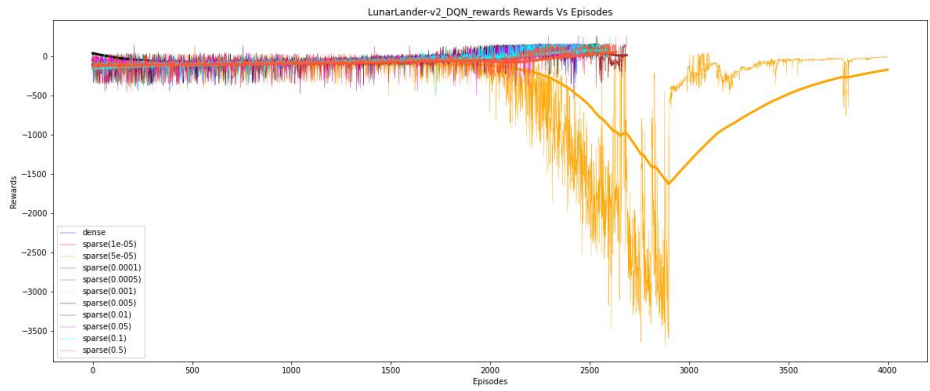

Figure 23: Reward Plots for the DQN policies in the LunarLander-v2 environment.

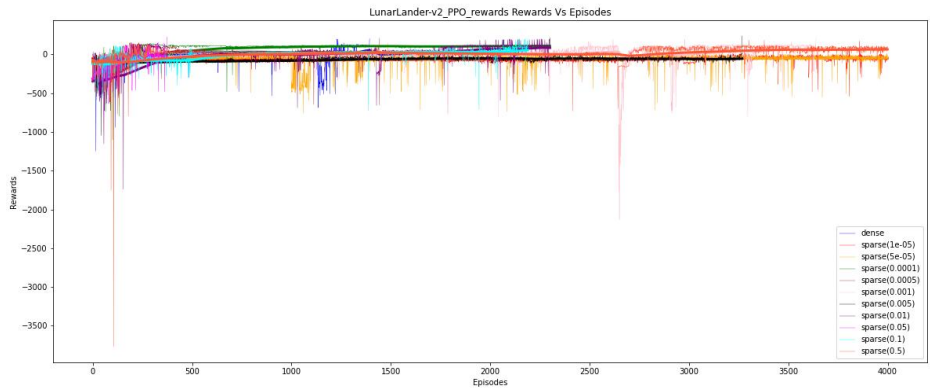

Figure 24: Reward Plots for the PPO policies in the LunarLander-v2 environment.

## A.2 Effect of Sparsification on Various Measures - Heatmap

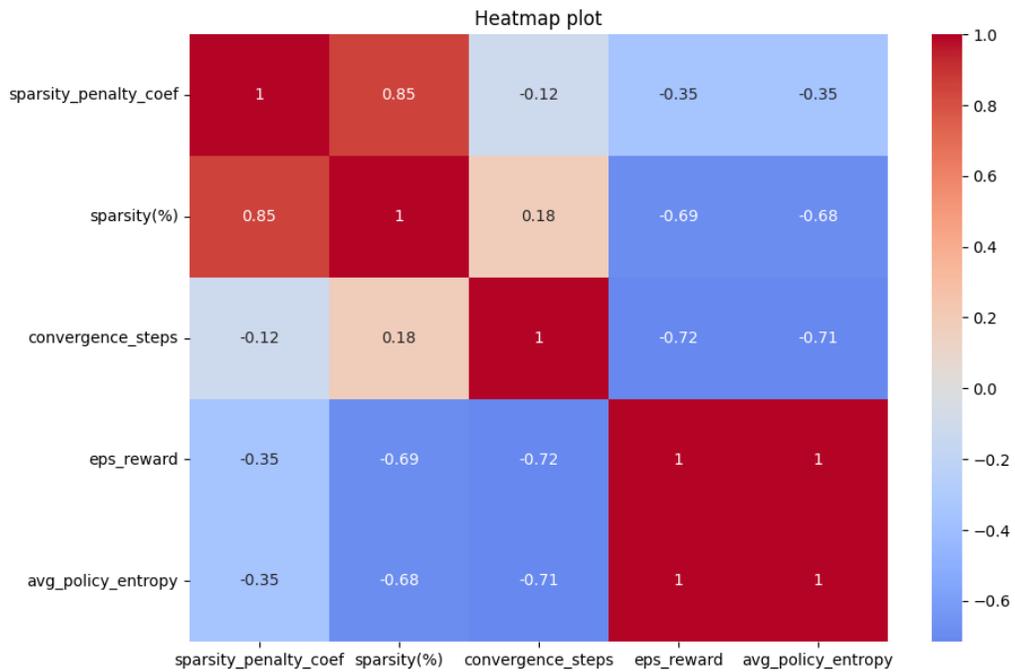

Figure 25: Benchmarking Results: Heatmap of the effects of sparsification on various measures of the DQN policies in the CartPole-v1 environment.

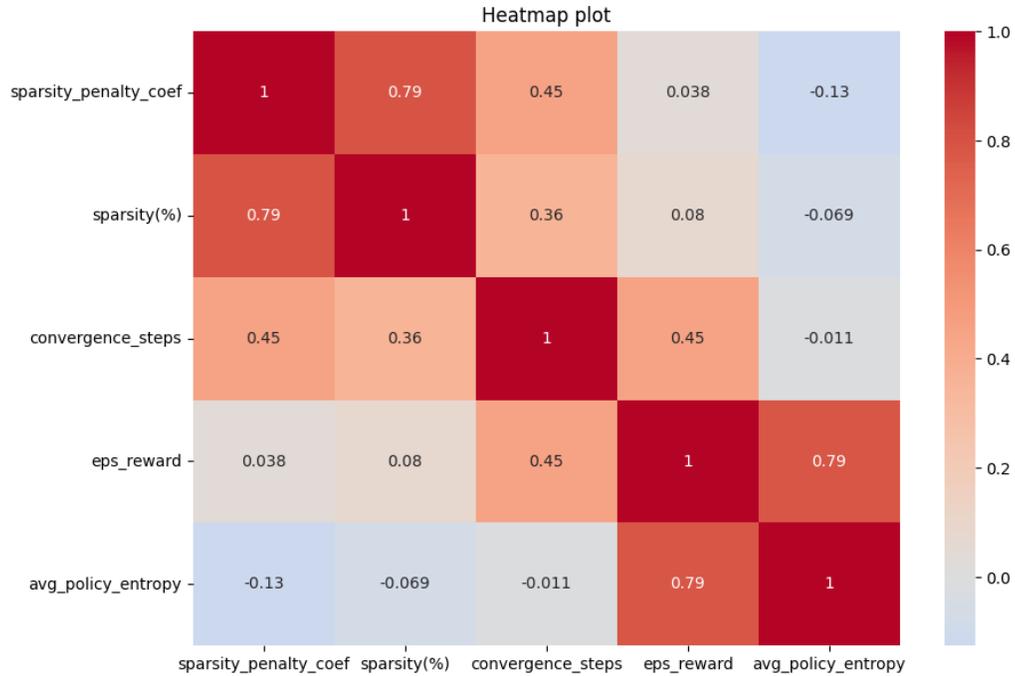

Figure 26: Benchmarking Results: Heatmap of the effects of sparsification on various measures of the PPO policies in the CartPole-v1 environment.

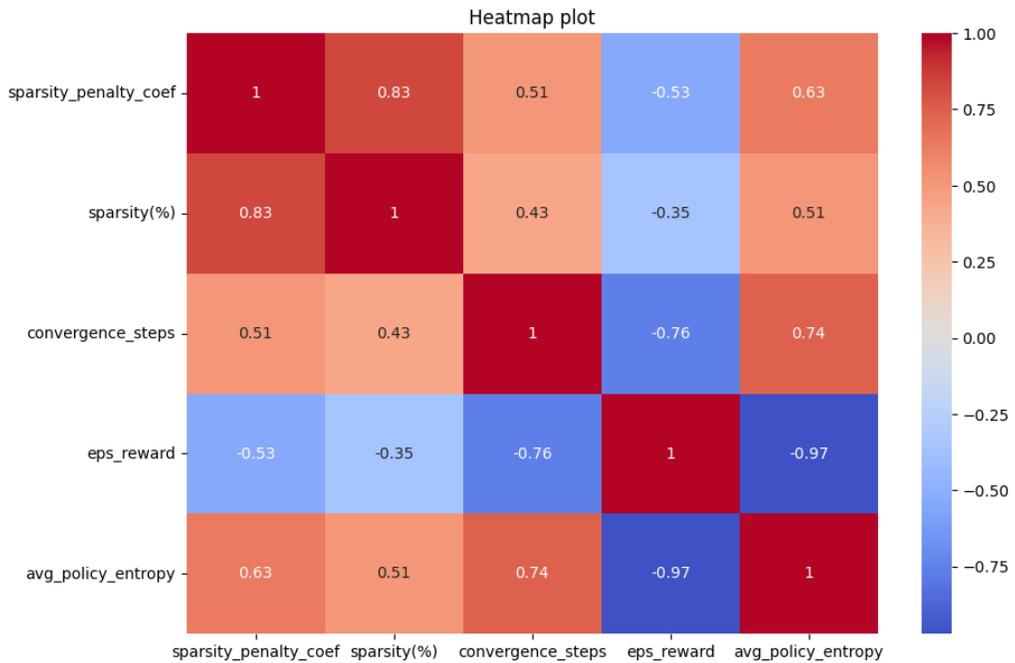

Figure 27: Benchmarking Results: Heatmap of the effects of sparsification on various measures of the DQN policies in the Acrobot-v1 environment.

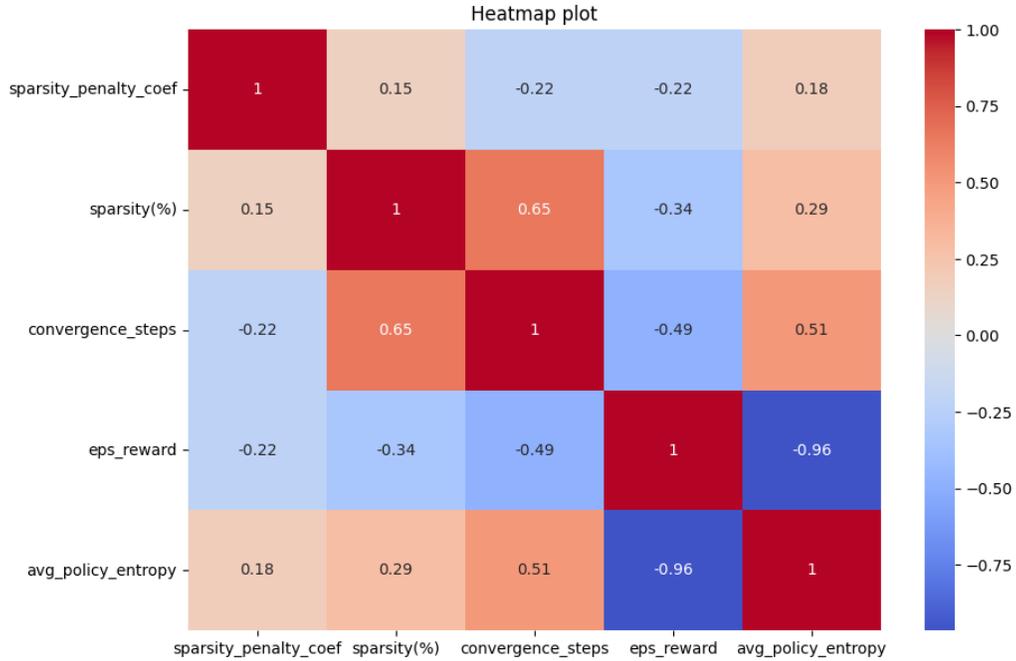

Figure 28: Benchmarking Results: Heatmap of the effects of sparsification on various measures of the PPO policies in the Acrobot-v1 environment.

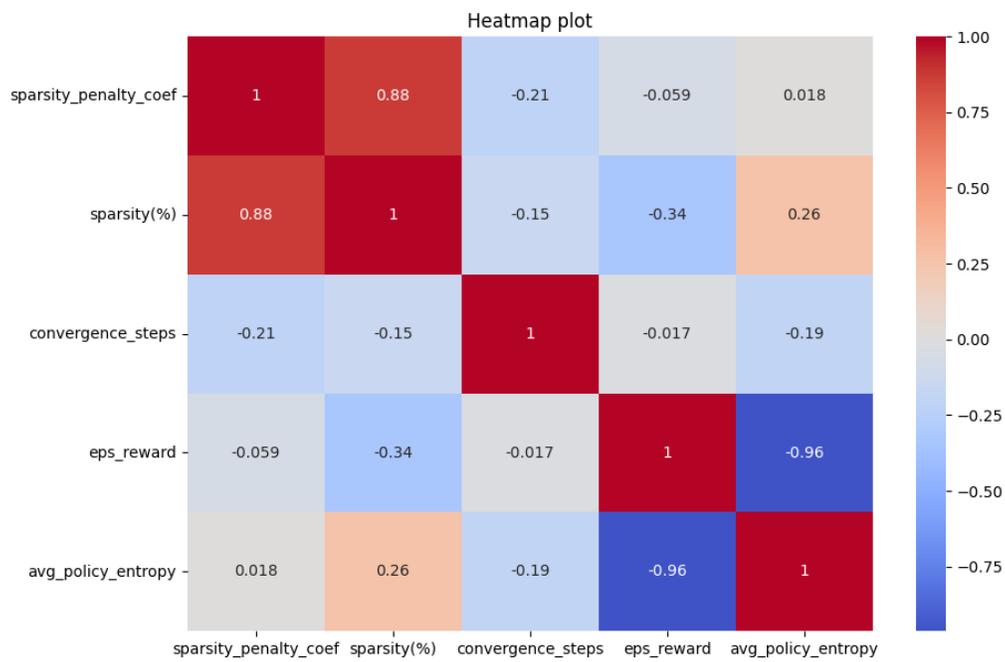

Figure 29: Benchmarking Results: Heatmap of the effects of sparsification on various measures of the DQN policies in the LunarLander-v2 environment.

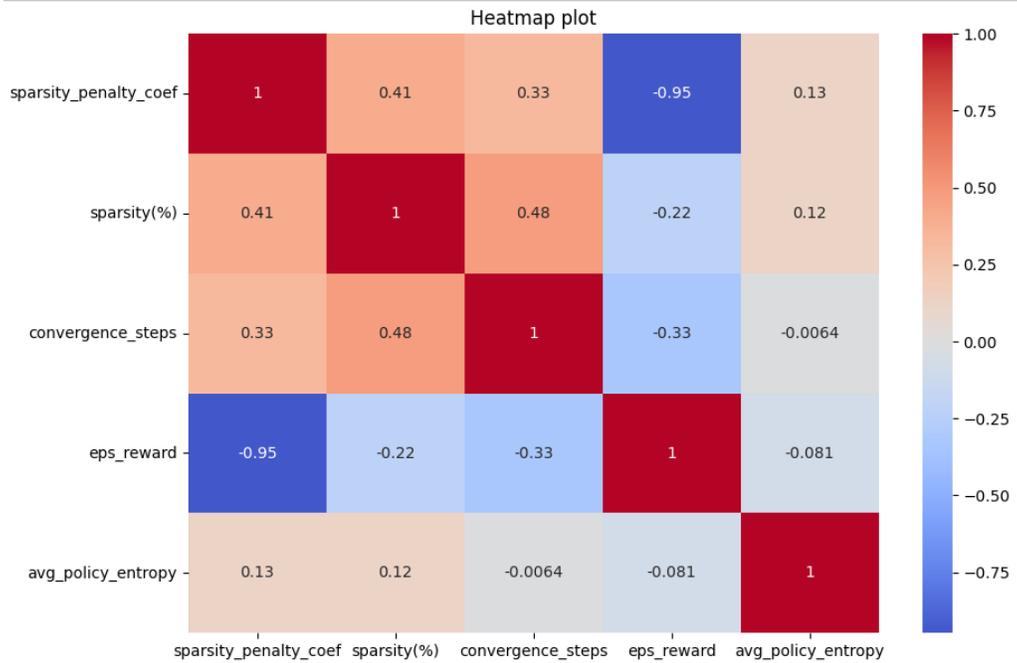

Figure 30: Benchmarking Results: Heatmap of the effects of sparsification on various measures of the PPO policies in the LunarLander-v2 environment.

Visually, we can infer that there seems to be no uniformity in the way the sparsification coefficients ($\lambda_{sp}$) affects the percentage sparsification, however it is worth noting that more often than not there seems to be strong positive correlations between both quantities, with the exception of *PPO* policies of the Acrobot-v1 and the LunarLander-v2 environments where weak positive correlation persists. Most of the other measures across the various environments and policies are not strongly affected by $\lambda_{sp}$, so we can arrive at the conclusion that the imposed effects by $\lambda_{sp}$ on them are random. Overall, the non-uniformity of the effect of $\lambda_{sp}$ on the percentage sparsification, as well as other measures can most likely be attributed to the stochastic nature of the batch gradient descent variant used for policy optimization.

## A.3 Sparsification of the PPO Policy in SuperMarioBros 7-1-v0 Environment

The following Figure illustrates the performance of dense and sparse PPO policies in the SuperMarioBros 7-1-v0 environment.

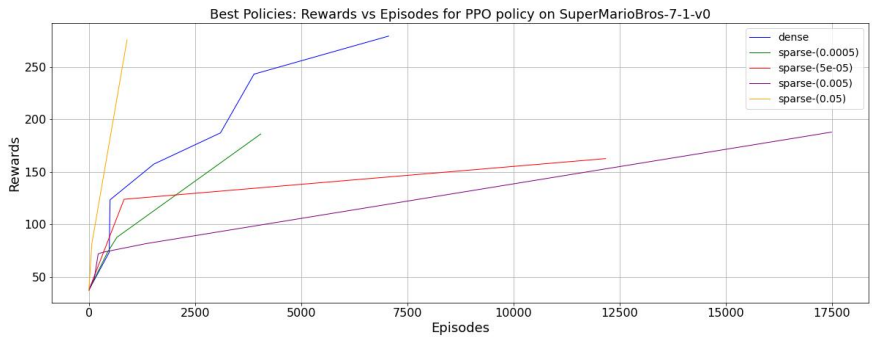

Figure 31: Effect of rank values on percentage policy size decrease.

## A.4 Low Rank Decomposition - Policy Size by Rank

In the following Figure, we outline the low rank decomposition and the corresponding decrease in policy size for Cartpole-V1, Acrobot-v1 and LunarLander-v2 environments.

| Rank | CartPole-v1 policies Size decrease (%) | Acrobot-v1 policies Size decrease (%) | LunarLander-v2 policies Size decrease (%) |
|---|---|---|---|
| 5 | 78.88 | 77.71 | 77.70 |
| 6 | 76.07 | 73.69 | 73.84 |
| 7 | 73.23 | 70.92 | 69.85 |
| 8 | 70.35 | 68.11 | 65.82 |
| 9 | 67.43 | 65.27 | 63.05 |
| 10 | 64.48 | 62.40 | 60.24 |
| 11 | 61.49 | 59.48 | 57.40 |
| 12 | 58.47 | 56.54 | 54.53 |
| 13 | 55.41 | 53.55 | 51.62 |
| 14 | 52.31 | 50.53 | 48.67 |
| 15 | 49.18 | 47.48 | 45.70 |
| 16 | 46.01 | 44.39 | 42.69 |
| 17 | 42.80 | 41.27 | 39.64 |
| 18 | 39.56 | 38.11 | 36.56 |
| 19 | 36.28 | 34.91 | 33.44 |
| 20 | 32.97 | 31.68 | 30.29 |
| 21 | 29.61 | 28.42 | 27.11 |
| 22 | 26.23 | 25.12 | 23.89 |
| 23 | 22.80 | 21.78 | 20.64 |
| 24 | 19.34 | 18.41 | 17.35 |
| 25 | 15.85 | 15.00 | 14.03 |
| 26 | 12.32 | 11.56 | 10.67 |
| 27 | 8.75 | 8.08 | 7.28 |
| 28 | 5.14 | 4.57 | 3.86 |
| 29 | 1.50 | 1.02 | 0.4 |

Figure 32: Effect of rank values on percentage policy size decrease.

In the following Figure, we outline the low rank decomposition and the corresponding performance of the dense and sparse DDQN policies in SuperMarioBros 7-1-v0 environment.

| Dense DDQN Policy | Rank | Eval Reward | % Size Decrease |
|---|---|---|---|
| | 10 | 79.75 | 95.62 |
| | 30 | 75.00 | 86.98 |
| | 50 | 81.00 | 78.48 |
| | 70 | 265.50 | 70.07 |
| | 90 | 265.50 | 61.93 |
| | 110 | 265.50 | 53.77 |
| | 130 | 279.00 | 45.56 |
| | 150 | 279.00 | 37.31 |
| | 170 | 37.00 | 29.02 |
| | 190 | 317.00 | 20.70 |
| Best Performing Sparse DDQN Policy ($\lambda_{sp} = 0.05$) | Rank | Eval Reward | % Size Decrease |
| | 10 | -25.00 | 95.62 |
| | 30 | -25.00 | 86.98 |
| | 50 | 197.00 | 78.48 |
| | 70 | 516.25 | 70.07 |
| | 90 | 516.25 | 61.93 |
| | 110 | 516.25 | 53.77 |
| | 130 | 516.25 | 45.56 |
| | 150 | 516.25 | 37.31 |
| | 170 | 516.25 | 29.02 |
| | 190 | 516.25 | 20.70 |

Figure 33: Low Rank Decomposition Results Table for Dense and best performing L0-norm sparse DDQN Policy.

## A.5 GitHub Links

The code-bse is organized at $https://github.com/vikramgoddla/RL_policy_sparsification$, which can be provided upon request.